\documentclass[11pt,a4paper]{article}
\usepackage[hyperref]{acl2019}
\usepackage{times}
\usepackage{latexsym}

\usepackage{url}
\usepackage{xspace}
\usepackage{siunitx}

\usepackage{graphicx}
\usepackage{multirow}
\usepackage{colortbl}
\usepackage{tabularx}
\usepackage{booktabs}
\usepackage{collcell}
\usepackage{amssymb}
\usepackage{mathtools,xparse}
\usepackage{csquotes}

\newcommand\ie{i.\,e.\xspace}
\newcommand\eg{e.\,g.\xspace}

\newcommand\etc{etc.\xspace}

\newcommand\cf{cf.\xspace}

\DeclarePairedDelimiter{\norm}{\lVert}{\rVert}

\aclfinalcopy 
\def\aclpaperid{1603} 

\newcommand\BkQA{RankQA\xspace}
\newcommand\Rr{\mathbb{R}}

\title{\BkQA: Neural Question Answering with Answer Re-Ranking}
\author{Bernhard Kratzwald$^\spadesuit$ \quad Anna Eigenmann$^\diamondsuit$ \quad Stefan Feuerriegel$^\spadesuit$ \\
	$^\spadesuit$ Chair of Management Information Systems, ETH Zurich \\
	$^\diamondsuit$ Department of Mathematics, ETH Zurich \\
	{\tt \{bkratzwald, eianna, sfeuerriegel\}@ethz.ch} \\
}

\begin{document}
\maketitle

\begin{abstract}
The conventional paradigm in neural question answering~(QA) for narrative content is limited to a two-stage process: first, relevant text passages are retrieved and, subsequently, a neural network for machine comprehension extracts the likeliest answer. However, both stages are largely isolated in the status quo and, hence, information from the two phases is never properly fused. In contrast, this work proposes \BkQA\footnote{Code is available from \url{https://github.com/bernhard2202/rankqa}}: \BkQA extends the conventional two-stage process in neural QA with a third stage that performs an additional answer re-ranking. The re-ranking leverages different features that are directly extracted from the QA pipeline, \ie, a combination of retrieval and comprehension features. While our intentionally simple design allows for an efficient, data-sparse estimation, it nevertheless outperforms more complex QA systems by a significant margin: in fact, \BkQA achieves state-of-the-art performance on 3 out of 4 benchmark datasets. Furthermore, its performance is especially superior in settings where the size of the corpus is dynamic. Here the answer re-ranking provides an effective remedy against the underlying noise-information trade-off due to a variable corpus size. As a consequence, \BkQA represents a novel, powerful, and thus challenging baseline for future research in content-based QA. 
\end{abstract}

\section{Introduction}
\begin{figure*}[h]
	\centering
	\includegraphics[width=.7\textwidth]{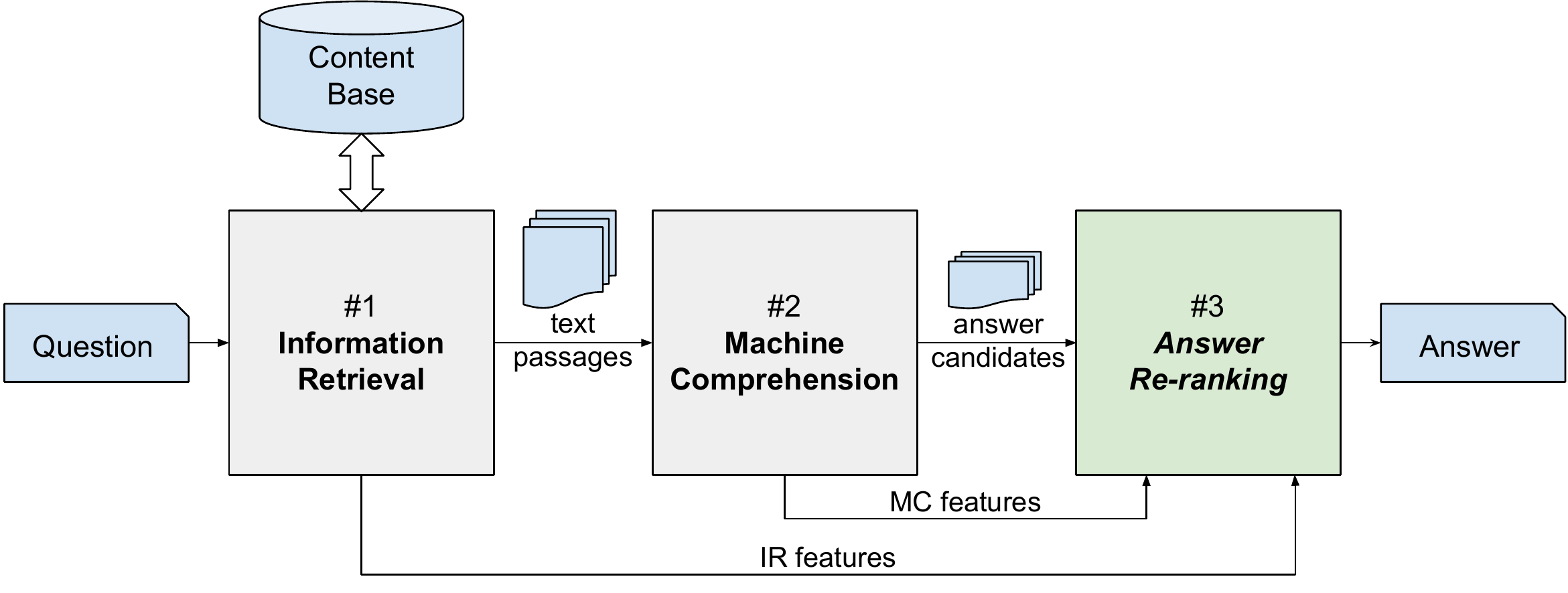}
	\caption{The \BkQA system consisting of three modules for information retrieval, machine comprehension, and our novel answer re-ranking. \BkQA fuses information from the information retrieval and machine comprehension phase to re-rank answer candidates within a full neural QA pipeline.}
	\label{fig:pipeline}
\end{figure*}


Question answering~(QA) has recently experienced considerable success in variety of benchmarks due to the development of neural QA \cite{Chen.2017, Wang.2018}. These systems largely follow a two-stage process. First, a module for information retrieval selects text passages which appear relevant to the query from the corpus. Second, a module for machine comprehension extracts the final answer, which is then returned to the user. This two-stage process is necessary for condensing the original corpus to passages and eventually answers; however, the dependence limits the extent to which information is passed on from one stage to the other. 


Extensive efforts have been made to facilitate better information flow between the two stages. These works primarily address the interface between the stages \cite{Lee.2018,Lin.2018}, \ie, which passages and how many of them are forwarded from information retrieval to machine comprehension. For instance, the QA performance is dependent on the corpus size and the number of top-$n$ passages that are fed into the module for machine comprehension \cite{Kratzwald.2018}. Nevertheless, machine comprehension in this approach makes use of only limited information (\eg, it ignores the confidence or similarity information computed during retrieval). 

State-of-the-art approaches for selecting better answers engineer additional features within the machine comprehension model with the implicit goal of considering information retrieval. For instance, the DrQA architecture of \citet{Chen.2017} includes features pertaining to the match between question words and words in the paragraph. Certain other works also incorporate a linear combination of paragraph and answer score \cite{Lee.2018}. Despite that, the use is limited to simplistic features and the potential gains of re-ranking remain untapped.

Prior literature has recently hinted at potential benefits from answer re-ranking, albeit in a different setting \cite{Wang.2017}: the authors studied multi-paragraph machine comprehension at sentence level, instead of a complete QA pipeline involving an actual information retrieval module over a full corpus of documents. However, when adapting it from a multi-paragraph setting to a complete corpus, this type of approach is known to become computationally infeasible \cite[\cf discussion in][]{Lee.2018}. In contrast, answer re-ranking as part of an actual QA pipeline not been previously studied.  


\textbf{Proposed \BkQA:} This paper proposes a novel paradigm for neural QA. That is, we augment the conventional two-staged process with an additional third stage for efficient answer re-ranking. This approach, named ``{\BkQA}'', overcomes the limitations of a two-stage process in the status quo whereby both stages operate largely in isolation and where information from the two is never properly fused. In contrast, our module for answer re-ranking fuses features that stem from both retrieval and comprehension. Our approach is intentionally light-weight, which contributes to an efficient estimation, even when directly integrated into the full QA pipeline. We show the robustness of our approach by demonstrating significant performance improvements over different QA pipelines. 


\textbf{Contributions:} To the best of our knowledge, \BkQA represents the first neural QA pipeline with an additional third stage for answer re-ranking. Despite the light-weight architecture, \BkQA achieves state-of-the-art performance across 3 established benchmark datasets. In fact, it even outperforms more complex approaches by a considerable margin. This particularly holds true when the corpus size is variable and where the resulting noise-information trade-off requires an effective remedy. Altogether, \BkQA yields a strong new baseline for content-based question answering. 

\section{\BkQA}

\BkQA is designed as a pipeline of three consecutive modules (see Fig.~\ref{fig:pipeline}), as detailed in the following. Our main contribution lies in the design of the answer re-ranking component and its integration into the full QA pipeline. In order to demonstrate the robustness of our approach, we later experiment with two implementations in which we vary module~2.

\subsection{Module 1: Information Retrieval}

For a given query, the information retrieval module retrieves the top-$n$ (here: $n = 10$) matching documents from the content repository and then splits these articles into paragraphs. These paragraphs are then passed on to the machine comprehension component. The information retrieval module is implemented analogously to the default specification of \citet{Chen.2017}, scoring documents by hashed bi-gram counts.

\subsection{Module 2: Machine Comprehension}

The machine comprehension module extracts and scores one candidate answer for every paragraph of all top-$n$ documents. Hence, this should result in $\gg n$ candidate answers; however, out of these, the machine comprehension module selects only the top-$k$ candidate answers $[c_1, \ldots, c_k]$, which are then passed on to the re-ranker. The size $k$ is a hyperparameter (here: $k = 40$). We choose two different implementations for the machine comprehension module in order to show the robustness of our approach.

\textbf{Implementation 1 (DrQA):} Our first implementation is based on the DrQA document reader \cite{Chen.2017}. This is the primary system in our experiments for two reasons. First, in neural QA, DrQA is a well-established baseline. Second, DrQA has become a widespread benchmark with several adaptations, which lets us compare our approach for answer re-ranking with other extensions that improve the retrieval of paragraphs \cite{Lee.2018} or limit the information flow between the retrieval and comprehension phases \cite{ Kratzwald.2018}.

\textbf{Implementation 2 (BERT-QA):} QA systems whose machine comprehension module is based on BERT are gaining in popularity~\cite{Yang.2019, Yang.2019a}. Following this, we implement a second QA pipeline where the document reader from DrQA is replaced with BERT~\cite{Devlin.2019}.\footnote{We used the official implementation from \url{https://github.com/google-research/bert}} We call this system BERT-QA and use it as a second robustness check in our experiments. 

\subsection{Module 3: Answer Re-Ranking}

Our re-ranking module receives the top-$k$ candidate answers $[c_1, \ldots, c_k]$ from the machine comprehension module as input. Each candidate $c_i$, $i = 1, \ldots, k$, consists of the actual answer span $s_i$ (\ie, the textual answer) and additional meta-information $\phi_i$ such as the document ID and paragraph ID from which it was extracted. Our module follows a three-step procedure in order to re-rank answers: 
\begin{description}
	\item [(i)~Feature extraction:] First, we extract a set of information retrieval and machine comprehension features for every answer candidate directly from the individual modules of the QA pipeline.
	\item[(ii)~Answer aggregation:] It is frequently the case that several answer candidates $c_i$ are duplicates and, hence, such identical answers are aggregated. This creates additional aggregation features, which should be highly informative and thus aid the subsequent re-ranking. 
	\item[(iii)~Re-ranking network:] Every top-$k$ answer candidate is re-ranked based on the features generated in (i) and (ii). 
\end{description}

\begin{table*}[h]
	\centering
	\footnotesize
	\begin{tabular}{p{3.5cm}p{6.5cm}ll}
		\toprule
		\bfseries Feature Group & \bfseries Description & \bfseries Aggregation  & \bfseries Impl.\\
		\midrule
		\multicolumn{4}{c}{\textsc{Information Retrieval Features}}\\
		\midrule
		Document-query similarity & Similarity between the question and the full document the answer was extracted from. & min, max, avg, sum& both\\
		Paragraph-query similarity & Similarity between the question and the paragraph the answer was extracted from. & ---& both\\
		Length features &Length of the document, length of the paragraph, and length of the question. &---& both\\
		Question type &The question type is a 13-dimensional vector indicating weather the questions started with the words \texttt{What was}, \texttt{What is}, \texttt{What}, \texttt{In what}, \texttt{In which}, \texttt{In}, \texttt{When}, \texttt{Where}, \texttt{Who}, \texttt{Why}, \texttt{Which}, \texttt{Is}, or \texttt{<other>}.& ---& both\\
		\midrule
		\multicolumn{4}{c}{\textsc{Machine Comprehension Features}}\\
		\midrule
		Span features & The score of the answer candidate as assigned directly from the MC  module, proportional to the probability of the answer given the paragraph, \ie, $\propto p(a|p)$.& min, max, avg, sum& both\\
		Named entity features& A 13-dimensional vector indicating whether one of following 13 named entities is contained within the answer span: \texttt{location}, \texttt{person}, \texttt{organization}, \texttt{money}, \texttt{percent}, \texttt{date}, \texttt{time}, \texttt{set}, \texttt{duration}, \texttt{number}, \texttt{ordinal}, \texttt{misc}, and \texttt{<other>}. & ---  & only 1\\
		Part-of-speech features  & A 45-dimensional vector indicating which part-of-speech tag is contained within the answer span. We use the Penn Treebank PoS tagset. \cite{Marcus.1993}. & --- & only 1 \\
		Ranking & Original ranking of the answer candidate. & number of occurrences& both\\
		\bottomrule
	\end{tabular} 
	\caption{Detailed description of all features used in our answer re-ranking component.}
		\label{tab:features}
\end{table*}

\subsubsection{Feature Extraction}

During this step, we extract several features from the information retrieval and machine comprehension modules for all top-$k$ answer candidates, which can later be fused; see a detailed overview in Tbl.~\ref{tab:features}. These features are analogously computed by most neural QA systems, albeit for other purposes than re-ranking. Nevertheless, this fact should highlight that such features can be obtained without additional costs. The actual set of features depends on the implementation of the QA system (\eg, DrQA extracts additional named entity features, as opposed to BERT-QA).

From the \emph{information retrieval} module, we obtain: (i)~the document-question similarity; (ii)~the paragraph-question similarity; (iii)~the paragraph length; (iv)~the question length; and (v)~indicator variables that specify with which word a question starts (\eg, ``what'', ``who'', ``when'', \etc).

From the \emph{machine comprehension} module, we extract: (i)~the original score of the answer candidate; (ii)~the original rank of the candidate answer; (iii)~part-of-speech tags of the answer; and (iv)~named entity features of the answer. The latter two are extracted only for DrQA and encoded via indicator variables that specify whether the answer span contains a named entity or part-of-speech tag (\eg, \texttt{PERSON}=1 or \texttt{NNS}=1).

\subsubsection{Answer Aggregation}

It is frequently the case that several candidate answers are identical and, hence, we encode this knowledge as a set of additional features. The idea of answer aggregation is similar to \citet{Lee.2018} and \citet{Wang.2017}, although there are methodological differences: the previous authors sum the probability scores for identical answers, whereas the aim in \BkQA is to generate a rich set of aggregation features.

That is, we group all answer candidates with an identical answer span. Formally, we merge two candidate answers $c_i$ and $c_j$ if their answer span is equal, \ie, $s_i=s_j$. We keep the information retrieval and machine comprehension features of the initially higher-ranked candidate $c_{\min \{ i,j \} }$. In addition, we generate further aggregation features as follows: (i)~the number of times a candidate with an equal answer span appears within the top-$k$ candidates; (ii)~the rank of its first occurrence; (iii)~the sum, mean, minimum, and maximum of the span scores; and (iv)~the sum, mean, minimum, and maximum of the document-question similarity scores. Altogether, this results, for each candidate answer $c_i$, in a vector $x_i$ containing all features from information retrieval, machine comprehension, and answer aggregation.

\subsubsection{Re-Ranking Network}

Let $x_i \in \Rr^d$ be the $d$-dimensional feature vector for the answer candidate $c_i$, $i = 1, \ldots, k$. We score each candidate via the following ranking network, \ie, a two-layer feed-forward network $f(x_i)$ that is given by
\begin{equation}
	f(x_i) = \mathrm{ReLU}(x_iA^T +b_1) \, B^T+b_2,
\label{eqn:nn}
\end{equation}
where $A \in \Rr^{m\times d}$ and  $B \in \Rr^{1\times m}$ are trainable weight matrices and where $b_1\in \Rr^m$ and $b_2\in \Rr$ are linear offset vectors.

During our experiments, we tested various ranking mechanisms, even more complicated architectures such as recurrent neural networks that read answers, paragraphs, and questions. Despite their additional complexity, the resulting performance improvements over our straightforward re-ranking mechanisms were only marginal and, oftentimes, we even observed a decline.

\subsection{Estimation: Custom Loss/Sub-Sampling}

The parameters in $f(\cdot)$ are not trivial to learn. We found that sampling negative (incorrect) and positive (correct) candidates, in combination with a binary classification loss or a regression loss, was not successful. As a remedy, we propose the following combination of ranking loss and sub-sampling, which proved beneficial in our experiments. 

We implement a loss $\mathcal{L}$, which represents a combination of a pair-wise ranking loss $\mathcal{L}_{\text{rank}}$ and an additional regularization $\mathcal{L}_{\text{reg}}$, in order to train our model. Given two candidate answers $i,j$ with $i \neq j$ for a given question, the binary variables $y_i$ and $y_j$ denote whether the respective candidate answers are correct or incorrect. Then we minimize the following pair-wise ranking loss adapted from \citet{Burges.2005}, \ie,
\begin{equation}
	\mathcal{L}_{\text{rank}}(x_i, x_j) = \bigg[ y_i - \sigma\left( f(x_i)-f(x_j) \right) \bigg]^2 .
\end{equation}
Here $f(\cdot)$ denotes our previous ranking network and $\sigma(\cdot)$ the sigmoid function. An additional penalty is used to regularize the parameters and prevent the network from overfitting. It is given by
\begin{equation}
	\mathcal{L}_{\text{reg}} = \norm{A}_1+\norm{B}_1+\norm{b_1}_1+\norm{b_2}_1 .
\end{equation} 
Finally, we optimize
	$\mathcal{L} = \mathcal{L}_{\text{rank}} + \lambda \mathcal{L}_{\text{reg}}$
 using mini-batch gradient descent with $\lambda$ as a tuning parameter.

We further implement a customized sub-sampling procedure, since the majority of candidate answers generated during training are likely to be incorrect. To address the pair-wise loss during sub-sampling, we proceed as follows: we first generate a list of answer candidates for every question in our training set using the feature extraction and aggregation mechanisms from our re-ranking. Then we iterate through this list and sample a pair of candidate answers $(x_i, x_j)$ if and only if they are at adjacent ranks ($i$ is ranked directly before $j$ \ie, iff $j=i+1$). We specifically let our training focus on pairs that are originally ranked high, \ie, $j<4$, and ignore training pairs ranked lower. During inference, we still score all top-$10$ answer candidates and select the best-scoring answer.
 
\section{Experimental Design}

\subsection{Content Base and Datasets}

Following earlier research, our content base comprises documents from the English Wikipedia. For comparison purposes, we use the same dump as in prior work \cite[\eg,][]{Chen.2017,Lee.2018}.\footnote{Downloaded from \url{https://github.com/facebookresearch/DrQA}} We do not use pre-selected documents or other textual content in order to answer questions.

We base our experiments on four well-established datasets. 
\begin{description}
	\item[SQuAD] The Stanford Question and Answer Dataset (SQuAD) contains more than \num{100000} question-answer-paragraph triples \cite{Rajpurkar.2016}. We use SQuAD\textsubscript{OPEN}, which ignores the paragraph information.
	\item[WikiMovies] This dataset contains several thousand question-answer pairs from the movie industry \cite{Miller.2016}. It is designed such that all questions can be answered by a knowledge-base (\ie, Open Movie Database) or full-text content (Wikipedia).
	\item[CuratedTREC] This dataset is a collection of question-answer pairs from four years of Text Retrieval Conference (TREC) QA challenges \cite{Baudis.2015}.
	\item[WebQuestions] The answers to questions in this dataset are entities in the Freebase knowledge-base \cite{Berant.2013}. We use the adapted version of \citet{Chen.2017}, who replaced the Freebase-IDs with textual answers.
\end{description}
\begin{table*}[h]
	\small
	\footnotesize
	\centering
	\begin{tabular}{l cccc}
		\toprule
		& {\bf SQuAD\textsubscript{OPEN}} & {\bf CuratedTREC} & {\bf WebQuestions} & {\bf WikiMovies}\\
		\midrule
		Baseline: DrQA~\cite{Chen.2017} & 29.8 & 25.4 & 20.7 & 36.5 \\
		\midrule
		\emph{DrQA extensions:} &&&&\\
		Paragraph Ranker~\cite{Lee.2018} & 30.2 & \bf35.4 & 19.9 & 39.1 \\
		Adaptive Retrieval& 29.6 & 29.3 & 19.6 & 38.4\\
		\cite{Kratzwald.2018}&&&&\\
		\midrule
		\emph{Other architectures:}&&&& \\
		$R^3$~\cite{Wang.2018} & 29.1 & 28.4 & 17.1 & 38.8 \\
		DS-QA~\cite{Lin.2018} &  --- & 29.1  & 18.5 & --- \\
		Min. Context~\cite{Min.2018} & \bf 34.6 & ---  & --- & --- \\
		\midrule 
		\BkQA (general) &34.5 & 32.4& \bf21.8 &\bf 43.3  \\
		\BkQA (task-specific) & \bf 35.3 &\bf 34.7 &\bf 22.3 &\bf 43.1\\ \midrule
		Upper bound: perfect re-ranking for $k=40$  &54.2 &65.9 &53.8 & 65.0 \\
		\bottomrule
	\end{tabular} 
	\caption{Exact matches of \BkQA compared to DrQA as natural baseline without re-ranking and state-of-the-art systems for neural QA. We use a general model that is trained on all datasets, and a task-specific model that is trained individually for every dataset. The two best results for every dataset are marked in bold.}
	\label{tab:main_results}
\end{table*}

\subsection{Training Details}

Our sourcecode and pre-trained model are available at: \url{https://github.com/bernhard2202/rankqa}.

\textbf{\BkQA:} The information retrieval module is based on the official implementation of \citet{Chen.2017}.\footnote{Available at \url{https://github.com/facebookresearch/DrQA}} The same holds true for the pre-trained DrQA-DS model, which we used without alterations. For BERT-QA, we use the uncased BERT base model and fine-tune it for three epochs on the SQuAD training split with the default parameters.\footnote{Available at \url{https://github.com/google-research/bert}} 

\textbf{Datasets:} We use the training splits of SQuAD, CuratedTREC, WikiMovies, and WebQuestions for training and model selection. In order to balance  differently-sized datasets, we use 10\,\% of the smallest training split for model selection and 90\,\% for training. For every other dataset, we take the same percentage of samples for model selection and all other samples for training.  We monitor the loss on the model selection data and stop training if it did not decrease within the last 10 epochs or after a total of 100 epochs. Finally, we use the model with the lowest error on the model selection data for evaluation. Analogous to prior work, we use the test splits of CuratedTREC, WikiMovies, and WebQuestions, as well as the development split for SQuAD, though only for the final evaluation. In order to account for different characteristics in the datasets, we train a task-specific model individually for every dataset following the same procedure.

\textbf{Parameters:} During training, we use Adam \cite{Kingma.2014} with a learning rate of $0.0005$ and a batch size of $256$. The hidden layer is set to $m=512$ units. We set the number of top-$n$ documents to $n = 10$ and the number of top-$k$ candidate answers that are initially generated to $k = 40$. We optimize $\lambda$ over $\lambda \in \{5 \cdot 10^{-4}, 5 \cdot 10^{-5} \}$. All numerical features are scaled to be within~$[0,1]$. Moreover, we apply an additional log-transformation. 

\section{Results}

We conduct a series of experiments to evaluate our \BkQA system. First, we evaluate the end-to-end performance over the four abovementioned benchmark datasets and compare our system to various other baselines. Second, we show the robustness of answer re-ranking by repeating these experiments with our second implementation, namely BERT-QA. Third, we replicate the experiments of \citet{Kratzwald.2018} to evaluate the robustness against varying corpus sizes. Fourth, we analyze errors and discuss feature importance in numerical experiments.

During our experiments, we measure the end-to-end performance of the entire QA pipeline in terms of exact matches. That is, we count the fraction of questions for which the provided answer matches one of the ground truth answers exactly. Unless explicitly mentioned otherwise, we refer to the first implementation, namely re-ranking based on the DrQA architecture.

\subsection{Performance Improvement from Answer Re-Ranking}

Tbl.~\ref{tab:main_results} compares performance across different neural QA systems from the literature. The DrQA system \cite{Chen.2017} is our main baseline as it resembles \BkQA without the answer re-ranking step. Furthermore, we compare ourselves against other extensions of the DrQA pipeline such as the \emph{Paragraph Ranker} \cite{Lee.2018} or \emph{Adaptive Retrieval} \cite{Kratzwald.2018}. Finally, we compare against other state-of-the-art QA pipelines, namely, \emph{$R^3$} \cite{Wang.2018}, \emph{DS-QA} \cite{Lin.2018}, and the \emph{Min. Context} system from \citet{Min.2018}. For \BkQA, we use, on the one hand, a general model that is trained on all four datasets simultaneously. On the other hand, we account for the different characteristics of the datasets and thus employ task-specific models that are trained separately on every dataset. 

A direct comparison between DrQA and \BkQA demonstrates a performance improvement from up to $7.0$ percentage points when using \BkQA, with an average gain of $4.9$ percentage points over all datasets. Given the identical implementation of information retrieval and machine comprehension, this increase is solely attributable to our answer re-ranking. Our \BkQA also outperforms all other state-of-the-art QA systems in 3 out of 4 datasets by a notable margin. This holds true for extensions of DrQA (Paragraph Ranker and Adaptive Retrieval) and other neural QA architectures ($R^3$ and DS-QA). 

This behavior is also observed in the case of the task-specific re-ranking model, which is trained for every dataset individually. Here we achieve performance improvements of up to $9.3$ percentage points, with an average performance gain of $5.8$ percentage points. The results on the CuratedTREC task deserve further discussion. Evidently, the dataset is particular in the sense that it is very sensitive to specific features. This is confirmed later in our analysis of feature importance and explains why the task-specific \BkQA is inferior the general model by a large margin. 

Finally, in the last row of Tbl.~\ref{tab:main_results}, we provide the results of a perfect re-ranker that always chooses the correct answer if present. This system represents an upper bound of the degree to which re-ranking could improve results without changing the information retrieval or machine comprehension models.

\subsection{Robustness Check: BERT-QA}
\begin{table*}[h]
	\small
	\footnotesize
	\centering
	\begin{tabular}{l cccc}
		\toprule
		& {\bf SQuAD\textsubscript{OPEN}} & {\bf CuratedTrec} & {\bf WebQuestions} & {\bf WikiMovies}\\
		\midrule
		Baseline: BERT-QA (no re-ranking)& 23.3 & 19.7 & 8.2 & 10.9  \\
		\BkQA (implementation 2)  & \bf 35.8 & \bf32.0 & \bf13.7 & \bf20.6 \\ \midrule
		Upper bound: perfect re-ranking for $k=40$ & 61.2 & 66.6 & 39.6 & 49.8 \\
		\bottomrule
	\end{tabular} 
	\caption{Exact matches of \BkQA based on the BERT-QA pipeline. We show results of the the pipline without re-ranking, the results obtained by our re-ranking model, and an upper bound (\ie, perfect re-ranking).}
	\label{tab:main_results2}
\end{table*}

In order to demonstrate the robustness of answer re-ranking across different implementations, we repeat experiments from above based on the BERT-QA system. The results are shown in Tbl.~\ref{tab:main_results2}. 

The first row displays the results without answer re-ranking. The second row shows the results after integrating our re-ranking module in the QA pipeline. As one can see, answer re-ranking yields significant performance improvements over all four datasets, ranging between $12.5$ and $5.5$ percentage points. The last row again lists an upper bound as would have been obtained by a perfect re-ranking system with access to the ground-truth labels. The performance differences between DrQA and BERT can be attributed to the fact that we trained BERT only on the SQuAD dataset, while the pre-trained DrQA model was trained on all four datasets. 

\begin{figure}[t]
	\centering
	\includegraphics[width=.40\textwidth]{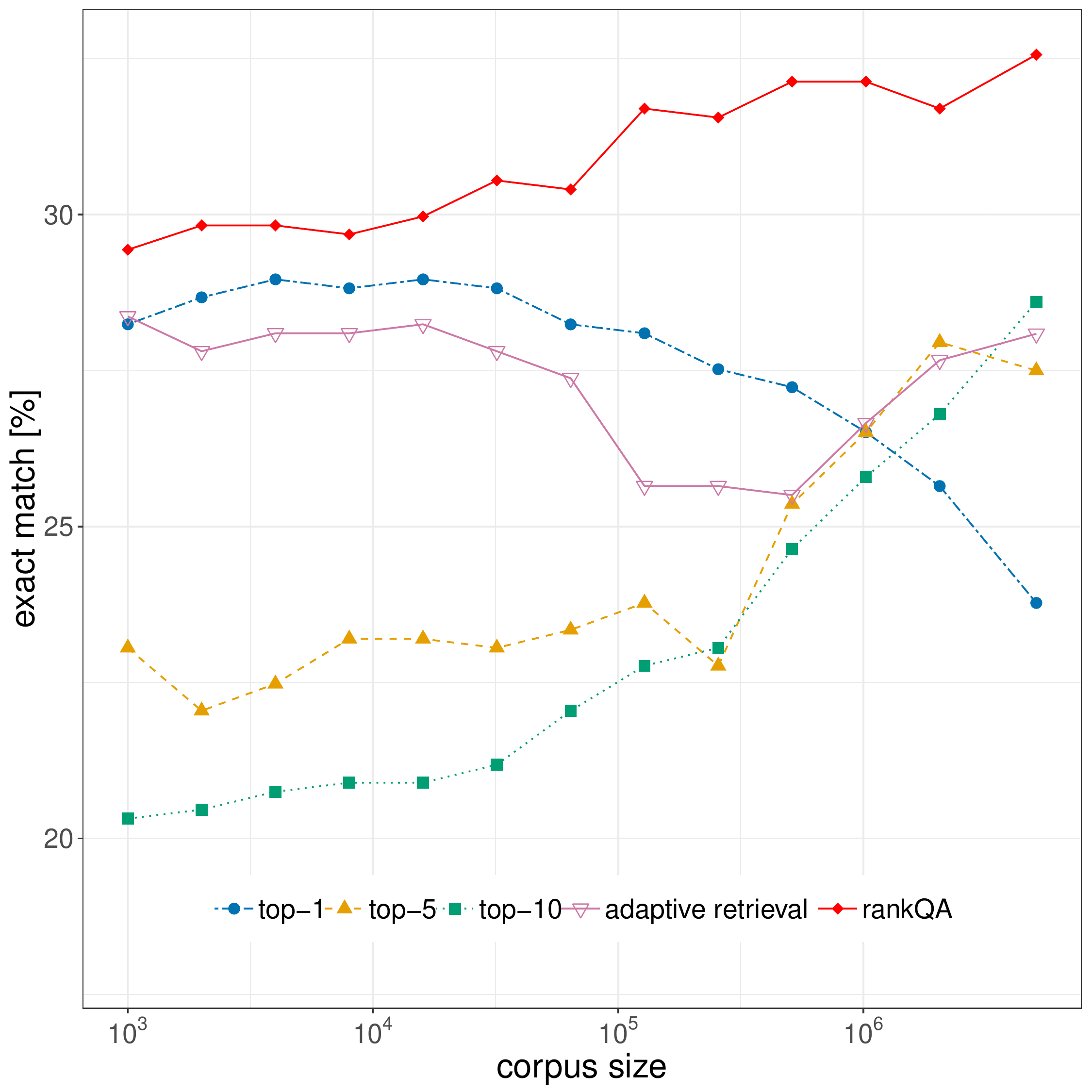}
	\caption{Robustness of answer re-ranking against a variable corpus size. We measure the exact matches for the CuratedTREC dataset while varying the corpus size from one thousand to over five  million documents.}
	\label{fig:adaptive_retrieval}
\end{figure}

\subsection{Performance Sensitivity to Corpus Size}

Corpora of variable size are known to pose difficulties for neural QA systems. \citet{Kratzwald.2018} ran a series of experiments in which they monitored the end-to-end performance of different top-$n$ systems (\ie, extracting the answer from the top-$10$ documents compared to extracting the answer from the top-$1$ document only). During the experiments, they increased the size of the corpus from one thousand to over five million documents. They found that selecting $n = 10$ is more beneficial for a large corpus, while $n = 1$ is preferable for small ones. They referred to this phenomenon as a noise-information trade-off: a large $n$ increases the probability that the correct answer is extracted, while a small $n$ reduces the chance that noisy answers will be included in the candidate list. As a remedy, the authors proposed an approach for \emph{adaptive retrieval} that chooses an independent top-$n$ retrieval for every query.

We replicated the experiments of \citet{Kratzwald.2018}\footnote{Source code for adaptive retrieval available at: \url{www.github.com/bernhard2202/adaptive-ir-for-qa}} and evaluated our \BkQA system in the same setting, as shown in Fig.~\ref{fig:adaptive_retrieval}. We see that answer re-ranking represents an efficient remedy against the noise-information trade-off. The performance of our system (solid red line) exceeds that of any other system configuration for any given corpus size. Furthermore, our approach behaves in a more stable fashion than adaptive retrieval. Adaptive retrieval, like many other recent advancements \cite[\eg,][]{Lee.2018,Lin.2018}, limits the amount of information that flows between the information retrieval and machine comprehension modules in order to select better answers. However, \BkQA does not limit the information, but directly re-ranks the answers to remove noisy candidates. Our experiments suggest that answer re-ranking is more  efficient than limiting the information flow when dealing with variable-size corpora.

\begin{table*}[h]
	\small
	\centering
	\begin{tabular}{l cccc}
		\toprule
		& {\bf SQuAD\textsubscript{OPEN}} & {\bf CuratedTrec} & {\bf WebQuestions} & {\bf WikiMovies}\\
		\midrule
		Baseline: DrQA~\cite{Chen.2017} & 29.8 & 25.4 & 20.7 & 36.5 \\
		\midrule
		\BkQA (general) &34.5 & 32.4& 21.8 & 43.3  \\
		\midrule
		\emph{Information Retrieval Features} &&&&\\
		\BkQA w/o query-document similarity &33.0&29.8&\underline{20.6}&42.0 \\
		\BkQA w/o query-paragraph similarity &32.1&32.0&22.0 & 42.1 \\
		\BkQA w/o length features &32.9&31.4&\bf22.3&42.6 \\
		\midrule
		\emph{Machine Comprehension Features} &&&&\\
		\BkQA w/o linguistic features (POS\&NER) &34.4&31.8&21.5&42.3 \\
		\BkQA w/o ranking features &34.1&31.8&21.4&43.3 \\
		\BkQA w/o span score &33.4&30.1&21.3&42.3 \\
		\midrule
		\emph{Feature Aggregation} &&&&\\
		\BkQA w/o aggregation features &33.6&26.9&\underline{18.5}&41.5\\
		\bottomrule
	\end{tabular}
	\caption{Feature importance (\ie, averaged performance of exact matches on a hold-out sample). We train the general model using the same data, but blind one group of features every time. We underline results that undershoot the baseline and mark results in bold that surpass the general model trained on all features. } 
	\label{tab:feature_importance}
\end{table*}

\subsection{Error Analysis and Feature Importance}
\label{sec:feature_importance}

We analyze whether our system is capable of keeping the set of correctly answered questions after applying the re-ranking step. Therefore, we measure the fraction of correctly answered questions out of those questions that had been answered correctly before re-ranking. Specifically, we found that the ratio of answers that remained correct varies between 94.6\,\% and 96.1\,\%. Hence, our model does not substantially change initially correct rankings.

\textbf{Feature importance:} Tbl.~\ref{tab:feature_importance} compares the relative importance of different features. This is measured by training the model with the same parameters and hyperparameters as before; however, we blind one (group of) feature(s) in every round. This was done as follows: when the information retrieval or machine comprehension features were blinded, we also removed the corresponding aggregated features. When omitting aggregation features, we keep the original un-aggregated feature. We show the performance of DrQA (\ie, system without answer re-ranking) and the full re-ranker for the sake of comparison. The original performance increase can only be achieved when all features are included. This has important implications for our approach to properly fusing information from information retrieval and machine comprehension. It suggests that aggregation features are especially informative and that it is not sufficient to use only a subset of those. 

We can see that individual datasets reveal a different sensitivity to all feature groups. The CuratedTREC or WebQuestions datasets, for instance, are highly sensitive to some information retrieval features. However, in all cases, the fused combination of features from both information retrieval and machine comprehension is crucial for obtaining a strong performance.  

\section{Related Work}

This work focus on question answering for unstructured textual content in English. Earlier systems of this type comprise various modules such as, for example, query reformulation \cite[\eg,][]{Brill.2002}, question classification~\cite{Li.2006}, passage retrieval~\cite[\eg,][]{Harabagiu.2000}, or answer extraction~\cite{Shen.2006}. However, the aforementioned modules have been reduced to two consecutive steps with the advent of neural QA.

\subsection{Neural Question Answering}

Neural QA systems, such as DrQA \cite{Chen.2017} or $R^3$ \cite{Wang.2018}, are usually designed as pipelines of two consecutive stages, namely a module for information retrieval and a module for machine comprehension. The overall performance depends on how many top-$n$ passages are fed into the module for machine comprehension, which then essentially generates multiple candidate answers out of which the one with the highest answer probability score is chosen. However, this gives rise to a noise-information trade-off \cite{Kratzwald.2018}. That is, selecting a large $n$ generates many candidate answers, but increases the probability of selecting the wrong final answer. Similarly, retrieving a small number of top-$n$ passages reduces the chance that the candidate answers contain the correct answer at all. 

Resolving the noise-information trade-off in neural QA has been primarily addressed by improving the interplay of modules for information retrieval and machine comprehension. \Citet{Min.2018} employ sentence-level retrieval in order to remove noisy content. Similarly, \citet{Lin.2018} utilize neural networks in order to filter noisy text passages, while \citet{Kratzwald.2018} forward a query-specific number of text passages. \Citet{Lee.2018} re-rank the paragraphs \emph{before} forwarding them to machine comprehension. However, none of the listed works introduce answer re-ranking to neural QA. 

\subsection{Answer Re-Ranking}

Answer re-ranking has been widely studied for systems other than neural QA, such as factoid~\cite{Severyn.2012}, non-factoid \cite{Moschitti.2011}, and definitional question answering \cite{Chen.2006}. These methods target traditional QA systems that construct answers in non-neural ways, \eg, based on $n$-gram tiling \cite{Brill.2002} or constituency trees \cite{Shen.2006}. However, neural QA extracts an answer directly from text using end-to-end trainable models, rather than constructing it. 

With respect to the conceptual idea, closest to our work is the approach of \citet{Wang.2017}, who use a single recurrent model to re-rank multiple candidate-answers given the paragraphs they have been extracted from. However, this work is different from our \BkQA in two ways. First, the authors must read multiple paragraphs in parallel via recurrent neural networks, which limits scalability and the maximum length of paragraphs; see the discussion in \citet{Lee.2018}. In contrast, our approach is highly scalable and can even be used together with complete corpora and long documents. Second, the authors evaluated their re-ranking in isolation, whereas we integrate our re-ranking into the full QA pipeline where the complete system is subject to extensive experiments. 

There are strong theoretical arguments as to why a better fusion of information retrieval and machine comprehension should be beneficial. First, features from information retrieval can potentially be decisive during answer selection (for instance, similarity features or document/paragraph length). Second, answer selection in state-of-the-art systems ignores linguistic features that are computed during the machine comprehension phase (\eg, DrQA uses part-of-speech and named entity information). Third, although some works aggregate scores for similar answers \cite[\eg,][]{Lee.2018,Wang.2017}, the complete body information is largely ignored during aggregation. This particularly pertains to, \eg, how often and with which original rank the top-$n$ answers were generated.

\section{Conclusion}

Our experiments confirm the effectiveness of a three-stage architecture in neural QA. Here answer re-ranking is responsible for bolstering the overall performance considerably: our \BkQA represents the state-of-the-art system for 3 out of 4 datasets. When comparing it to corresponding two-staged architecture, answer re-ranking can be credited with an average performance improvement of $4.9$ percentage points. This performance was even rendered possible with a light-weight architecture that allows for the efficient fusion of information retrieval and machine comprehension features during training. Altogether, \BkQA provides a new, strong baseline for future research on neural QA.  

\section*{Acknowledgments}
We thank the anonymous reviewers for their helpful comments. We gratefully acknowledge the support of NVIDIA Corporation with the donation of the Titan Xp GPUs used for this research.

\bibliographystyle{acl_natbib}
\bibliography{acl2019_long}

\end{document}